\pdfoutput=1

\documentclass[11pt]{article}

\usepackage[]{ACL2023}

\usepackage{times}
\usepackage{latexsym}

\usepackage[T1]{fontenc}

\usepackage[utf8]{inputenc}

\usepackage{microtype}

\usepackage{inconsolata}

\usepackage{diagbox}
\usepackage{mathrsfs}
\usepackage{nicematrix}
\usepackage{multirow}
\usepackage{booktabs}
\usepackage{graphicx}
\usepackage{subfigure}
\usepackage{bbm}
\usepackage{makecell}
\PassOptionsToPackage{prologue,dvipsnames}{xcolor}
\usepackage[dvipsnames]{xcolor}
\definecolor{myorange}{RGB}{237, 125, 48}
\definecolor{myblue}{RGB}{0, 113, 191}

\title{More than Classification: \\ A Unified Framework for Event Temporal Relation Extraction}

\author{
    Quzhe Huang$^{1,2}$, 
    Yutong Hu$^{1,2}$, 
    Shengqi Zhu$^3$,
    \\
    \textbf{Yansong Feng}$^{1}$\thanks{\;\;Corresponding author.}~~,  
    \textbf{Chang Liu}$^{1,4}$, 
    \textbf{Dongyan Zhao}$^{1,5}$ \\
    $^1$Wangxuan Institute of Computer Technology, Peking University, China \\ 
    $^2$School of Intelligence Science and Technology, Peking University \\
    $^3$University of Washington \\
    $^4$Center for Data Science, Peking University \\
    $^5$National Key Laboratory of General Artificial Intelligence \\
    {\tt \{huangquzhe,huyutong,fengyansong,liuchang97,zhaody\}} 
     {\tt @pku.edu.cn} \\
     {\tt sqzhu@uw.edu}\\
}

\begin{document}
\maketitle
\begin{abstract}
Event temporal relation extraction~(ETRE) is usually formulated as a multi-label classification task, where each type of relation is simply treated as a one-hot label. This formulation ignores the meaning of relations and wipes out their intrinsic dependency. After examining the relation definitions in various ETRE tasks, we observe that all relations can be interpreted using the start and end time points of events. For example, relation \textit{Includes} could be interpreted as event 1 starting no later than event 2 and ending no earlier than event 2. In this paper, we propose a unified event temporal relation extraction framework, which transforms temporal relations into logical expressions of time points and completes the ETRE by predicting the relations between certain time point pairs. Experiments on TB-Dense and MATRES show significant improvements over a strong baseline and outperform the state-of-the-art model by 0.3\% on both datasets. By representing all relations in a unified framework, we can leverage the relations with sufficient data to assist the learning of other relations, thus achieving stable improvement in low-data scenarios. When the relation definitions are changed, our method can quickly adapt to the new ones by simply modifying the logic expressions that map time points to new event relations. The code is released at \url{https://github.com/AndrewZhe/A-Unified-Framework-for-ETRE}.

\end{abstract}

\section{Introduction}
\label{sec:intro}

In order to fully understand natural language utterances, it is important to understand the temporal information conveyed in the text, especially the relations between the events~\cite{pustejovsky2003timeml,pustejovsky2010iso}. Such temporal relations play an essential role in downstream applications, such as question answering, event timeline generation, and information retrieval~\cite{choubey-huang-2017-sequential,han2019joint}. The Event Temporal Relation Extraction (ETRE) task is proposed to address the extraction of temporal relations between event pairs from text.

\begin{figure}[pt]
\center
  \includegraphics[width=0.49\textwidth]{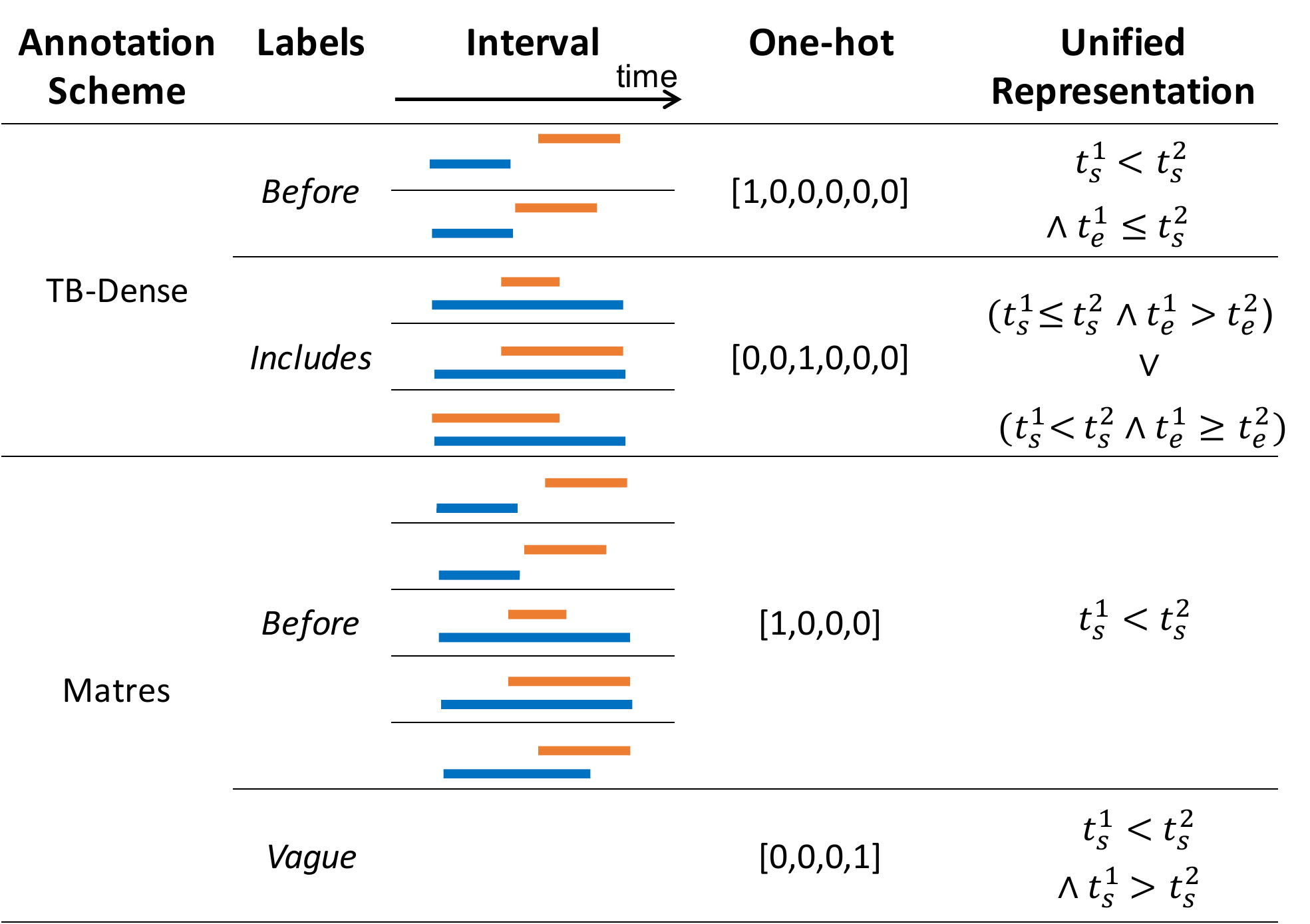}
    \caption{Examples of labels from TB-Dense and MATRES and their Interval, One-hot and Unified representations. \textcolor{myblue}{\rule[1.5pt]{0.5cm}{0.15em}} and \textcolor{myorange}{\rule[1.5pt]{0.5cm}{0.15em}} represent the intervals of event 1 and event 2 in the timeline. $t^*_s$ and $t^*_e$ represent the start and end time points of an event.
    }
    \label{fig:intro}
\end{figure}
Researchers have different ideas on how to define the temporal relations between two events. \citet{allen1981interval} treats an event as an interval in the timeline and uses 13 relations between two intervals to define the temporal relations between events. The 13 relations, together with a special relation \textit{Vague}, are then adopted by TimeBank \cite{pustejovsky2003timebank}. However, such a definition is so fine-grained that some relations are very hard to distinguish from each other. Thus following works make a simplification and aggregate some relations \cite{uzzaman2013semeval, styler-iv-etal-2014-temporal}. For example, TB-Dense \cite{cassidy2014annotation} aggregates \textit{Before} and \textit{Before Immediately} into one coarse relation \textit{Before}. Other studies, like MATRES \cite{ning2018multi}, think that identifying the duration of events requires very long contexts and even commonsense, making it exceedingly difficult to determine when an event ends. Therefore, in MATRES, only the start time of events is considered for temporal relations. 
We show some examples of temporal relations and their interval representations in Figure~\ref{fig:intro}.
It can be seen that despite the differences across definitions, each relation reflects certain aspects of the full temporal relationship and has rich meanings behind a single label.

Although the meaning of a relation is important, previous studies did not pay enough attention. They solve ETRE as a simple text classification task, first using an encoder to get the event pair representation and then feeding it into a multi-layer perceptron to get the prediction. All efforts are focused on generating better event pair representations, such as pre-training a task-specific language model \cite{han2020econet} or applying Graph Neural Networks to incorporate syntactic information \cite{DBLP:journals/corr/abs-2104-09570}.  However, the relations are only used as one-hot labels to provide guidance in cross-entropy loss. Such a classification-based method cannot fully use the meaning of relations and could cause the following problems:

\textbf{Misunderstanding Relations:} Some relations may correspond to complex scenarios, such as  \textit{Vague} in MATRES. It describes a contradictory situation where event 1 may occur before event 2 and event 2 may also occur before event 1. Such complex meaning cannot be conveyed by a simple one-hot vector.

\textbf{Missing the Dependency:} The classification-based method treats different relations as orthogonal vectors, however, relations within the same task definition are not independent. For example, both the relations \textit{Includes} and \textit{Before} in TB-Dense imply that event 1 does not start later than event 2.

\textbf{Lacking Generalization:} Since there is no one-to-one mapping between different relation definitions, the classification-based method needs training a unique classifier for every definition. For example, the relation \textit{Includes} in TB-Dense contains three interval relations, and only two of them overlap with relation \textit{Before} in MATRES. Therefore, when a classifier trained on TB-Dense predicts \textit{Includes}, it cannot figure out which relation it should predict under the definition of MATRES.

To address the aforementioned issues, we need a unified framework that can interpret any single relation and connect different ones. We go back to Allen's interval theory, and notice that the relation between intervals is determined by their endpoints, which represent the start and end time points of events. As nearly all definitions of ETRE are based on Allen's interval representation, we find that we can use the relation among the start and end time points of events to represent the relations in any definitions. As illustrated in Figure~\ref{fig:intro}, \textit{Includes} in TB-Dense could be represented as $(t_s^1 \leq t_s^2 \wedge t_e^1 > t_e^2) \vee (t_s^1 < t_s^2 \wedge t_e^1 \geq t_e^2) $. 

Inspired by this finding, we design a unified temporal relation extraction framework based on the time points of events. Specifically, based on the relation definitions, we first transform each relation into a logical expression of time point relations, as shown in the last column of Figure~\ref{fig:intro}. Then the task of predicting the temporal relation between events becomes the task of predicting the relation of time points. Following the annotation guidelines by \citet{ning2018multi}, we infer the relation between two time points t1 and t2 by asking the model two questions: 1) whether t1 could occur earlier than t2 and 2) whether t2 could occur earlier than t1. By answering these questions, we can deepen the association of different time point relations.

Our experiments show that the unified framework can significantly help temporal relation extraction, compared to a strong baseline, and outperforms state-of-the-art (SOTA) model by 0.3\% F1 on both TB-Dense and MATRES. By using time points to explicitly interpret the relations, we help the model to better understand ambiguous relations such as \textit{Vague}, and significantly reduce the number of instances misclassified as \textit{Vague}. In addition, since different relations can all be represented as logic expressions of the same time points, we can capture the dependency between different relations. The relations with more training data can be used to assist the learning of relations with fewer data, thus achieving stable improvement in low-data scenarios. When the definitions of temporal relations are changed, we can easily adapt to the new ones by modifying the logic expressions that map time points to new event relations. Further experiments with ChatGPT\footnote{https://chat.openai.com/} show that our unified framework can also help Large Language Models(LLMs), outperforming classification-based prompts by 2.3\% F1 on TB-Dense.

\section{Problem Formulation}
Given an input sequence $\mathbf{X}$ with two events $e_1$ and $e_2$, the task of event temporal relation extraction is to predict a relation from $\mathcal{R} \ \cup$ \{\textit{Vague}\} between the event pair ($e_1$ and $e_2$), where $\mathcal{R}$ is a pre-defined set of temporal relations of interests. Label \textit{Vague} means the relation between the two events can not be determined by the given context.

\section{Enhanced Baseline Model}
We first introduce our baseline model for ETRE. It is based on a strong entity relation extraction model~\cite{zhong2021frustratingly} and we integrate other techniques to make it suitable in ETRE. Our baseline can achieve comparable or even better performance with the previous SOTA in ETRE, providing a powerful encoder for our unified framework.

\subsection{Event Encoder}
\label{sec:event_encoder}
Given two event mentions $(e_1, e_2)$ and a text sequence $\mathbf{X} = [x_1, x_2, ..., x_n]$ of $n$ tokens, the event encoder aims to calculate the representation of the event pair. Considering cross-sentence information has been proven useful in entity relation extraction~\cite{DBLP:conf/emnlp/WaddenWLH19}, we believe ETRE will also benefit from it. Thus we extend the input text with 1 more sentence from the left and right context of the sentence containing mentions. To highlight the event mentions, we insert event markers <EVENT\_1>, </EVENT\_1>, <EVENT\_2> and </EVENT\_2> into the sequence $\mathbf{X}$ before and after the two events. The new sequence with text markers inserted is then fed into a pre-trained language model, and we use the contextual embeddings of <EVENT\_1> and <EVENT\_2>, denoted as $\mathbf{h}_{e_1}$ and $\mathbf{h}_{e_2}$ respectively, to calculate the representation of the event pair: 

\begin{equation}
    \mathbf{ee} = [\mathbf{h}_{e_1} \oplus \mathbf{h}_{e_2} ]
    \label{eq:event_pair}
\end{equation}
where $[* \oplus *]$ is the concatenation operator.

\subsection{Classifier}
Following previous efforts in ETRE~\cite{wen-ji-2021-utilizing,DBLP:journals/corr/abs-2104-09570}, our baseline uses a multi-layer perceptron~(MLP) and a softmax layer to convert the representation of the event pair into a probability distribution:
\begin{equation}
    P(\mathbf{R}|e_1, e_2) = \operatorname{softmax}(\mathbf{MLP}(ee))
\end{equation}
where $P(\mathbf{R_i}|e_1, e_2)$ denotes the probability of relation $i$ exiting between $e_1$ and $e_2$. We use the cross entropy loss for training.

\subsection{Label Symmetry}
Inspired by \citet{DBLP:journals/corr/abs-2104-09570,hwang-etal-2022-event}, based on the symmetry property of temporal relations, we expand our training set using rules provided in Appendix~\ref{App:sym_rule} while keeping the validation set and test set unchanged.

\section{Our Unified Framework}
\begin{figure}[pt]
\center
  \includegraphics[width=0.49\textwidth]{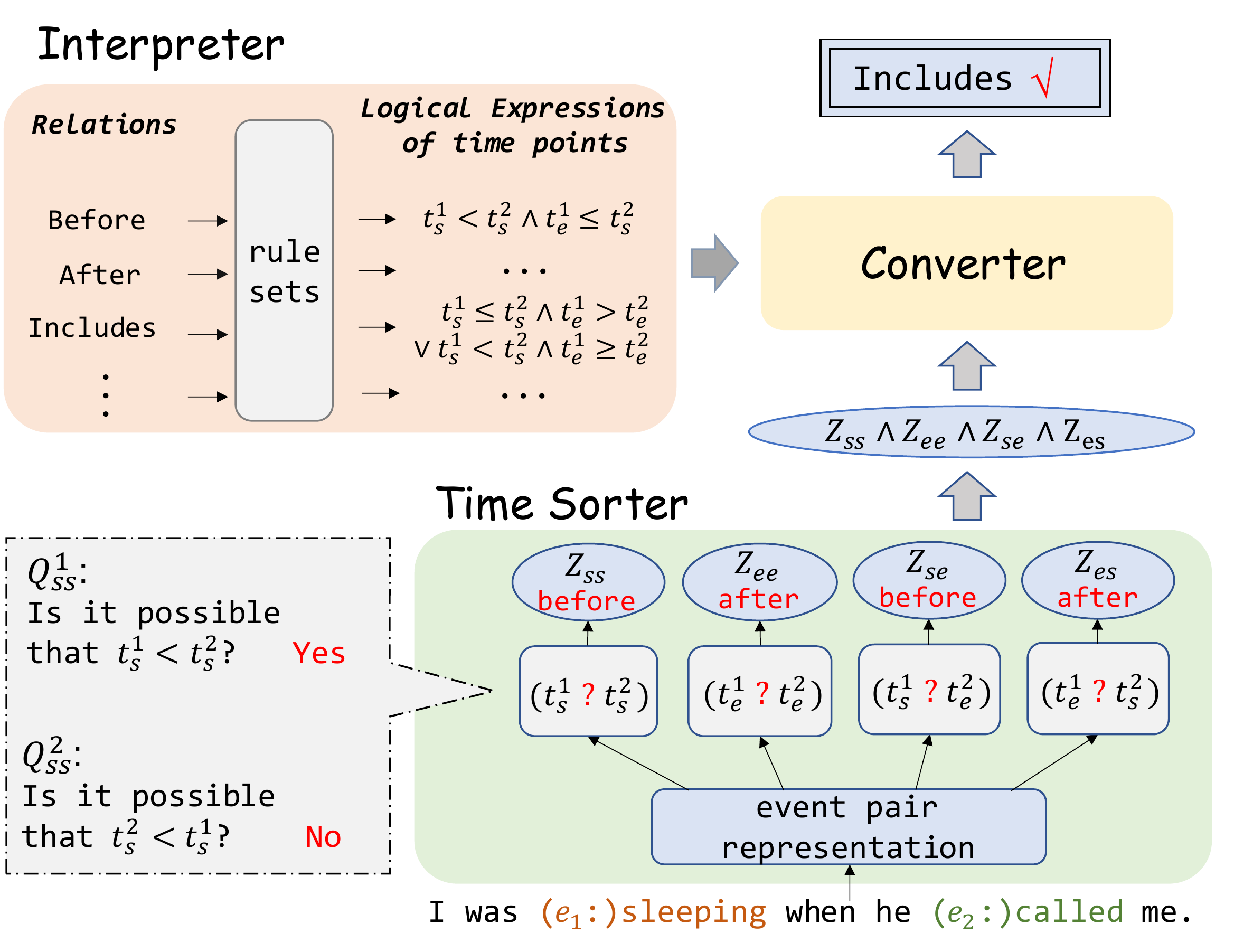}
    \caption{Architecture Overview}
    \label{fig:model}
\end{figure}
Generally, going through the classifier, a model can easily output a probability for each category. However, it is difficult for a model to understand what a category represents from the one-hot formed supervision signals, and the model will struggle in summarizing the category's meaning from training data.
Meanwhile, since the categories, which actually are relations, are treated as orthogonal signals for supervision, the data of a particular relation cannot help the model understand other relations.

To help the model make better use of the temporal information embedded in relations, we transform the task of predicting the temporal relations into a judgment of the relationship between the start and end time points of two events, which are the basic elements that make up temporal relations in different ETRE definitions.

As shown in Figure~\ref{fig:model},  our unified framework is composed of three parts: the first is the interpreter, which translates each relation into a logical expression of time points; the second part is the temporal predictor, which predicts the relation between time points based on the representation of an event pair; finally, the converter checks which logical expression is satisfied with the assignments from the second stage and thus infer the relations between two events.

\begin{figure}[pt]
\center
  \includegraphics[width=0.48\textwidth]{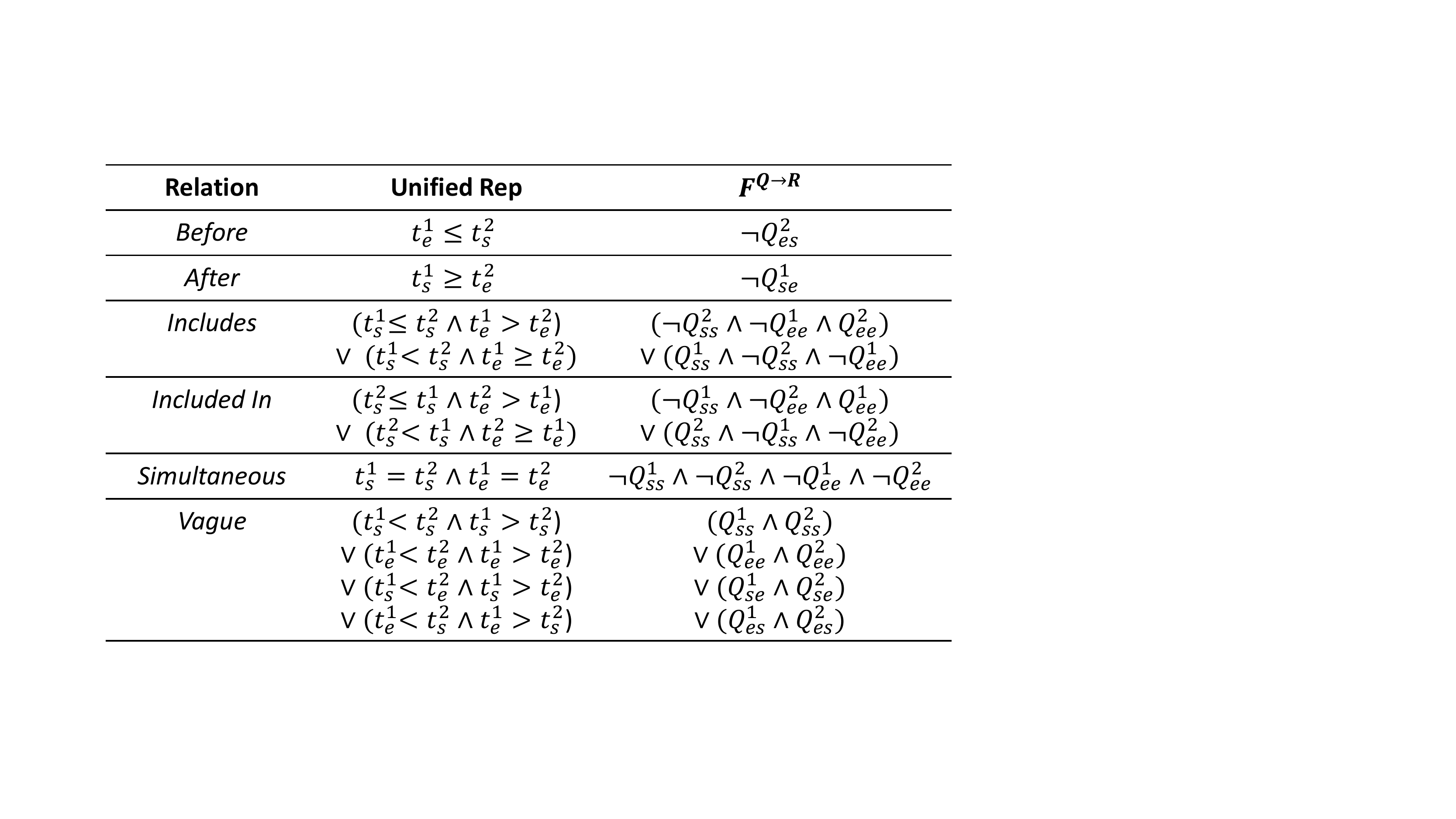}
    \caption{Relations in TB-Dense, and their unified representations and the logical expressions from Q to R($F^{Q\rightarrow}R$)}
    \label{fig:TB-Dense-relations}
\end{figure}
\subsection{Interpreter}
Following Allen's theory, the events $e_1$ and $e_2$ could be represented as two intervals $[t^1_s, t^1_e]$ and $[t^2_s, t^2_e]$, where $t^*_s$ and $t^*_t$ are the start and end times of an event. The event temporal relation then could be represented as the relation between intervals, which is determined by the endpoints of intervals, $t^1_s, t^1_e, t^2_s$ and $t^2_e$, for example, the interval relation \textit{Before} could be represented as $t^1_s < t^1_e < t^2_s < t^2_e$. Considering the start time of an event should not be later than its end time, to infer the interval's relation, we only need to consider the relations between four  pairs of time points,  which are $Z_{ss}(t^1_s, t^2_s), Z_{ee}(t^1_e, t^2_e), Z_{se}(t^1_s, t^2_e)$ and $Z_{es}(t^1_e, t^2_s)$. We show all the 13 interval relations and their time point representations in Appendix \ref{App:allen_relations}.

Current definitions of temporal relations, such as TB-Dense and Matres, are built up by aggregating some interval relations into one to form a coarse-grained relation set. As they are based on Allen's interval theory, we can also use the time points of $t^1_s, t^1_e, t^2_s$, and $t^2_e$ to represent their coarse-grained relations. For example, the relation \textit{Includes} in TB-Dense could be interpreted as $(t^1_s \leq t^2_s \wedge t^1_e > t^2_e) \vee (t^1_s < t^2_s \wedge t^1_e \geq t^2_e)$. 

The interpreter contains a set of rules to transform the definition of relations into the logical expressions of start and end time points of events. Figure~\ref{fig:TB-Dense-relations} shows the logical expressions of every relation in TB-Dense\footnote{We shows that of MATRES in Appendix \ref{App:matres-relations}}. The logical expressions of different relations are guaranteed to be mutually exclusive, as long as the two relations do not overlap with each other.

\subsection{Time Point Sorter}
\label{sec:time_soter}
There are four relations between two time points, \textit{before}, \textit{after}, \textit{equal}, and \textit{vague}. We could treat it as a four-label classification and use a MLP and a softmax layer to complete the prediction. However, such a method also treats each relation as orthogonal labels and cannot interpret the complex relation, \textit{vague}. Inspired by the annotation guidance in MATRES~\cite{ning2018multi}, we ask the model to answer the following two questions to decide the relation between two time points $t^1$ and $t^2$: \underline{$Q^1$: Is it possible that $t^1$ occur earlier than $t^2$?} and \underline{$Q^2$: Is it possible that $t^2$ occur earlier than $t^1$?}

The model only needs to answer \textit{yes} or \textit{no} to these two questions and the time point relation could be inferred by the rules in Table~\ref{tab:qzrules}.

\begin{table}[h]
\center
\begin{tabular}{lllll}
\hline
$Q^1$ & yes    & no    & no    & yes \\
$Q^2$ & no     & yes   & no    & yes \\ \hline
Z  & \textit{before} & \textit{after} & \textit{equal} & \textit{vague}  \\ \hline
\end{tabular}
\caption{Mapping from Q to Z}\label{tab:qzrules}
\end{table}

On the one hand, it makes a clear definition of relations like \textit{vague}, which helps the model understand such relations. On the other hand, the dependency between time point relations could be reflected in the same answer to one question, e.g., $Q^2$ for both relation \textit{before} and \textit{equal} is \textit{no}, which means it is impossible that $t^2$ is earlier than $t^1$ in both of these two relations.

To obtain the answers for $\mathbf{Q}$’s, we use a two-layer perceptron to simulate the procedure of answering the questions:
\begin{equation}
    logit^i_{tp} = FFN^2_{tp}(\sigma(FFN^1_{tp}(\mathbf{ee})))
\end{equation}
\begin{equation}
    P(Q^i_{tp}) = \operatorname{sigmoid}(\frac{logit^i_{tp}}{\tau}) 
    \label{eq:P_Q}
\end{equation}
\begin{equation}
    Q^i_{tp} = \mathbbm{1} \{P(Q^i_{tp}) > 0.5 \}
    \label{eq:Q}
\end{equation}
where time point pair $tp\in\{ss, ee, se, es\}$, $i\in\{1, 2\}$,  $\mathbbm{1}$ denotes the indicator function, $\operatorname{sigmoid}(\frac{*}{\tau})$ is a sigmoid function with temperature $\tau$ used to control the smoothing degree of the probability distribution, $P(Q^i_{tp})$ denotes the probability of answering \textit{yes} for the question $i$ of time point pair $tp$ and $Q^i_{tp}$ is the binary answer, 1 for \textit{yes} and 0 for \textit{no}.

\subsection{Converter}
After predicting the value of $\mathbf{Z}$, that is, we have obtained the relations between the start and end time points of two events, we need to check which logical expression in the interpreter is True under this set of assignments. As relations are exclusive to each other, we will find only one logical expression with True value and the relation corresponding to this expression will be the temporal relation between the events.

\subsection{Inference}
As discussed above, the mapping from $\mathbf{Q}$ to $\mathbf{Z}$ and the mapping from $\mathbf{Z}$ to $\mathbf{R}$ could both be represented as logical expressions. Thus, we could also use a logical expression of $\mathbf{Q}$ to directly represent the relations between events, which is denoted as $F^{\mathbf{Q}\rightarrow \mathbf{R}}$. Table~\ref{fig:TB-Dense-relations} shows the logical expressions of all relations in TB-Dense\footnote{The relations of MATRES are shown in Appendix~\ref{App:matres-relations}}.

\subsection{Training with Soft Logic}
\label{sec:soft_logic}
So far, we have discussed how to use hard logic to infer the event relation $\mathbf{R}$ from the values of $\mathbf{Q}$. However, in practice, the hard logic reasoning procedure is not differentiable. 
We thus use soft logic \cite{bach2017hinge} to encode the logic expressions from $\mathbf{Q}$ to $\mathbf{R}$. Specifically, soft logic allows continuous truth values from the interval [0, 1] instead of \{0,1\} \cite{hu2016harnessing}, and the Boolean logic operators are reformulated as:
\begin{align*}
    & a \land b = a \cdot b \\
    & a \lor b = a + b - a \cdot b \\
    & \neg a = 1 - a
\end{align*}
where $\land$ and $\lor$ are approximations to logical conjunction and disjunction.

We substitute the Boolean operators in $F^{\mathbf{Q}\rightarrow \mathbf{R}}$ with soft logic operators to get a differentiable mapping from Q to R, $F^{\mathbf{Q}\rightarrow \mathbf{R}}_{soft}$, and the probability of \textbf{R} can be formed as:
\begin{equation}
    P(\mathbf{R}) = F^{\mathbf{Q}\rightarrow \mathbf{R}}_{soft}(P(\mathbf{Q}))
\end{equation}
\begin{equation*}
    P(\mathbf{Q}) = \{P(Q_{tp}^i)| \\ tp \in \{ss, ee, se, es\}, i \in \{1, 2\}\}
\end{equation*}
where $P(Q_{tp}^i)$ is calculated by Equation~\ref{eq:P_Q}. With the probability of $\mathbf{R}$, we can use the normal cross-entropy loss function to train our model.

\section{Experiments}

\paragraph{Datasets}

We conduct experiments over two temporal relation extraction benchmarks, TB-Dense \cite{cassidy2014annotation} and MATRES \cite{ning2018multi}, both of which could be used for research purposes. 
TB-Dense includes 6 types of temporal relations:  \textit{Before}, \textit{After}, \textit{Includes}, \textit{Is\_Included}, \textit{Simultaneous} and \textit{Vague}. Temporal relations in MATRES are annotated only based on start time points, reducing them to 4 types:  \textit{Before}, \textit{After}, \textit{Equal} and \textit{Vague}. we use the same train/dev/test splits as previous studies \cite{han2019joint,wen-ji-2021-utilizing}. 

\paragraph{Evaluation Metrics} For fair comparisons with previous research, we adopt the same evaluation metrics as \citet{DBLP:journals/corr/abs-2104-09570}. On both TB-Dense and MATRES, we exclude the \textit{Vague} label and compute the micro-F1 score of all the others.

\subsection{Main Results}
\begin{table*}[t]
    \centering
    \small
    \begin{tabular}{c|c|c|c}
        \toprule
        \toprule
        \bf Model & \bf Pretrained Model & \bf TB-Dense & \bf MATRES\\
        \midrule
        LSTM \citep{cheng-miyao-2017-classifying} & BERT-Base & 62.2 &  73.4 \\
        CogCompTime2.0 \citep{ning-etal-2019-improved} & BERT-Base & - & 71.4 \\
        HNP \citep{han2019joint} & BERT-Base & 64.5 &  75.5 \\
        Box \citep{hwang-etal-2022-event}& RoBERTa-Base & - & 77.3 \\
        Syntactic \citep{DBLP:journals/corr/abs-2104-09570}* & BERT-Base & 66.7 & 79.3 \\
        \midrule
        Enhanced-Baseline & BERT-Base & 64.8 $\pm$ 0.85 & 78.5 $\pm$ 0.69 \\
        Unified-Framework (Ours) & BERT-Base & 66.4 $\pm$ 0.40 & 79.3 $\pm$ 0.45 \\
        \midrule
        \midrule
        PSL \citep{Zhou2021ClinicalTR} & RoBERTa-Large & 65.2 & - \\
        HMHD \citep{wang2021joint}& RoBERTa-Large & - & 78.8 \\
        DEER \citep{han2020econet}& RoBERTa-Large & 66.8 & 79.3 \\
        Time-Enhanced \citep{wen-ji-2021-utilizing}& RoBERTa-Large & - & 81.7\\
        HGRU \citep{tan2021extracting}& RoBERTa-Large & - & 80.5 \\
        Syntactic
        \citep{DBLP:journals/corr/abs-2104-09570}* & BERT-Large & 67.1 & 80.3 \\
        SCS-EERE
        \citep{Man_Ngo_Van_Nguyen_2022} & RoBERTa-Large & - & 81.6 \\
        TIMERS
        \citep{mathur-etal-2021-timers} & BERT-Large & 67.8 & 82.3 \\
        \midrule
        Enhanced-Baseline & RoBERTa-Large & 66.2 $\pm$ 2.08 & 81.9 $\pm$ 0.35 \\
        Unified-Framework (Ours) & RoBERTa-Large & 68.1 $\pm$ 1.35 & 82.6 $\pm$ 1.05 \\
        \bottomrule
        \bottomrule 
    \end{tabular}
    
    \caption{F1 score on TB-Dense and MATRES. Models marked * use additional training resources. SCS-EERE only reports the maximum score among multi-experiments. We re-run the code provided in the original work and report the average F1 among 3 experiments here.}
    \label{table:tre_compare}
\end{table*}

Table \ref{table:tre_compare} reports the performance of our unified framework and baseline methods on TB-Dense and MATRES. Overall, applying our unified  framework brings significant improvements compared with Enhanced-Baseline, and outperforms previous SOTA by 0.3\% F1 on both datasets, respectively. The only difference between ours and Enhanced-Baseline is that we use the events' start and end points to infer the relation and Enhanced-Baseline directly predicts the relation. The stable improvements on both benchmarks indicate our unified framework could help the model better understand temporal relations. 

Compared to the improvements on MATRES, which are 0.8\% and 0.7\% for BERT-Base and RoBERTa-Large respectively, we have a more significant gain on TB-Dense, which is nearly 2\% for both base and large models. This is because MATRES only cares about the start time of events and thus cannot benefit from our interpreter module. The improvements on MATRES show that in time point sorter, splitting the decision of time point relations into answering $Q_1$ and $Q_2$ is effective. And the greater improvements in TB-Dense further illustrate the usefulness of the interpreter module.

In addition, we notice consistent gains with either BERT-Base or RoBERTa-Large as the backbone. On TB-Dense, our method outperforms Enhanced-Baseline with about 2\% F1 for both BERT-Base and RoBERTa-Large, and the gain is about 1\% for both these two backbones on MATRES. The consistent improvement implies that the efficacy of our unified framework is orthogonal to the encoders' capability. We evaluate our methods with a very strong encoder, whose baseline version is comparable with the SOTA in MATRES, to show the benefits of using a unified framework will not disappear with the development of strong encoders. We believe, in the future, with better event pair representations, e.g., incorporating syntactic information like \citet{DBLP:journals/corr/abs-2104-09570}, our framework would remain effective.

\section{Analysis}
We further explore how our module makes better use of label information in ETRE tasks. We show the 3 problems of classification-based methods,  mentioned in Introduction, could be alleviated by our unified framework.

\subsection{Better Comprehension of Relations}
\begin{figure}[t]
    \centering
    \subfigure[TB-Dense]{
        \label{fig:subfig:tbd}
        \includegraphics[width=3.7cm, height=2.7cm]{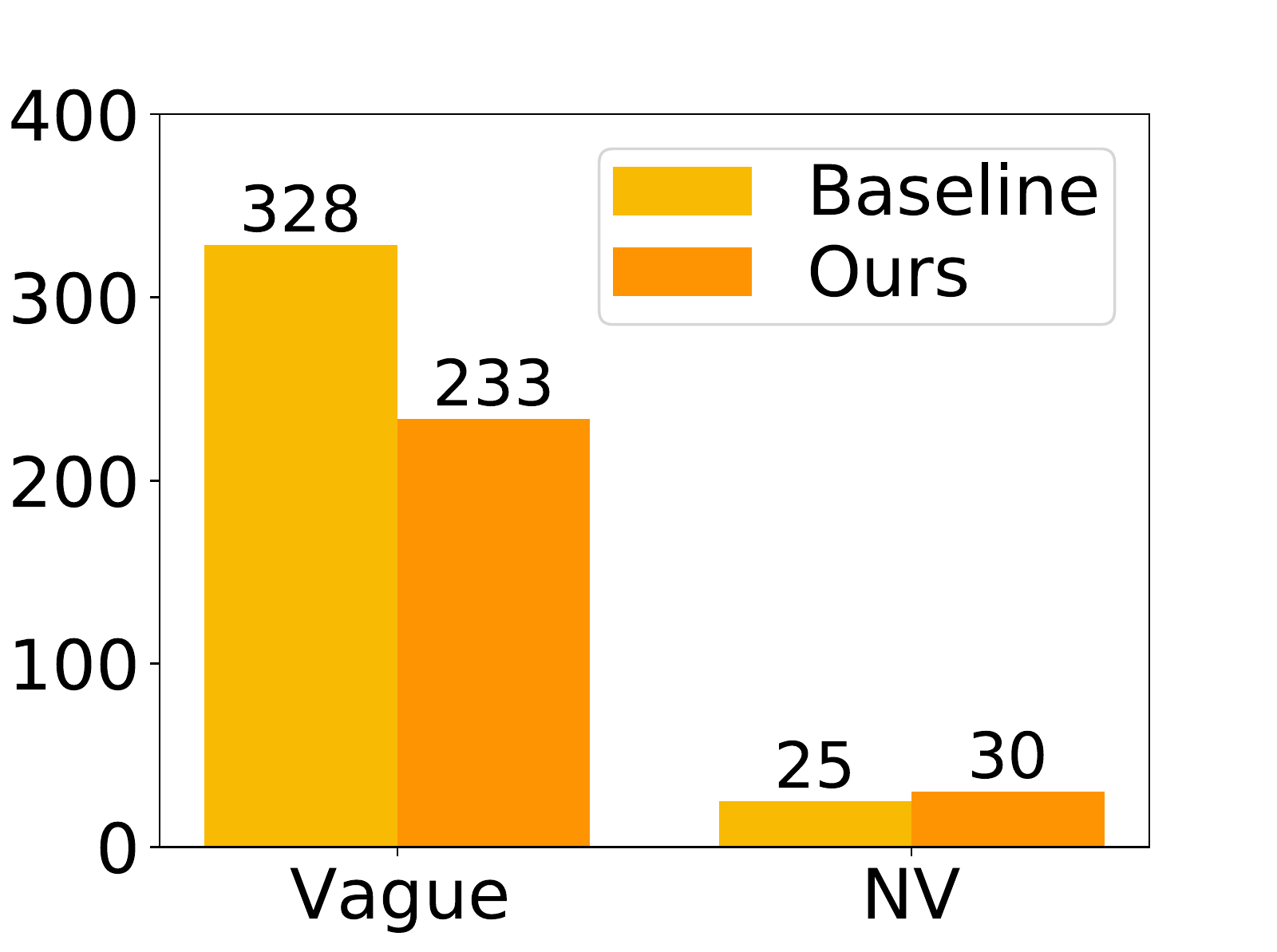}
        }
    \hspace{-2mm}
    \subfigure[MATRES]{
        \label{fig:subfig:mtr}
        \includegraphics[width=3.6cm, height=2.7cm]{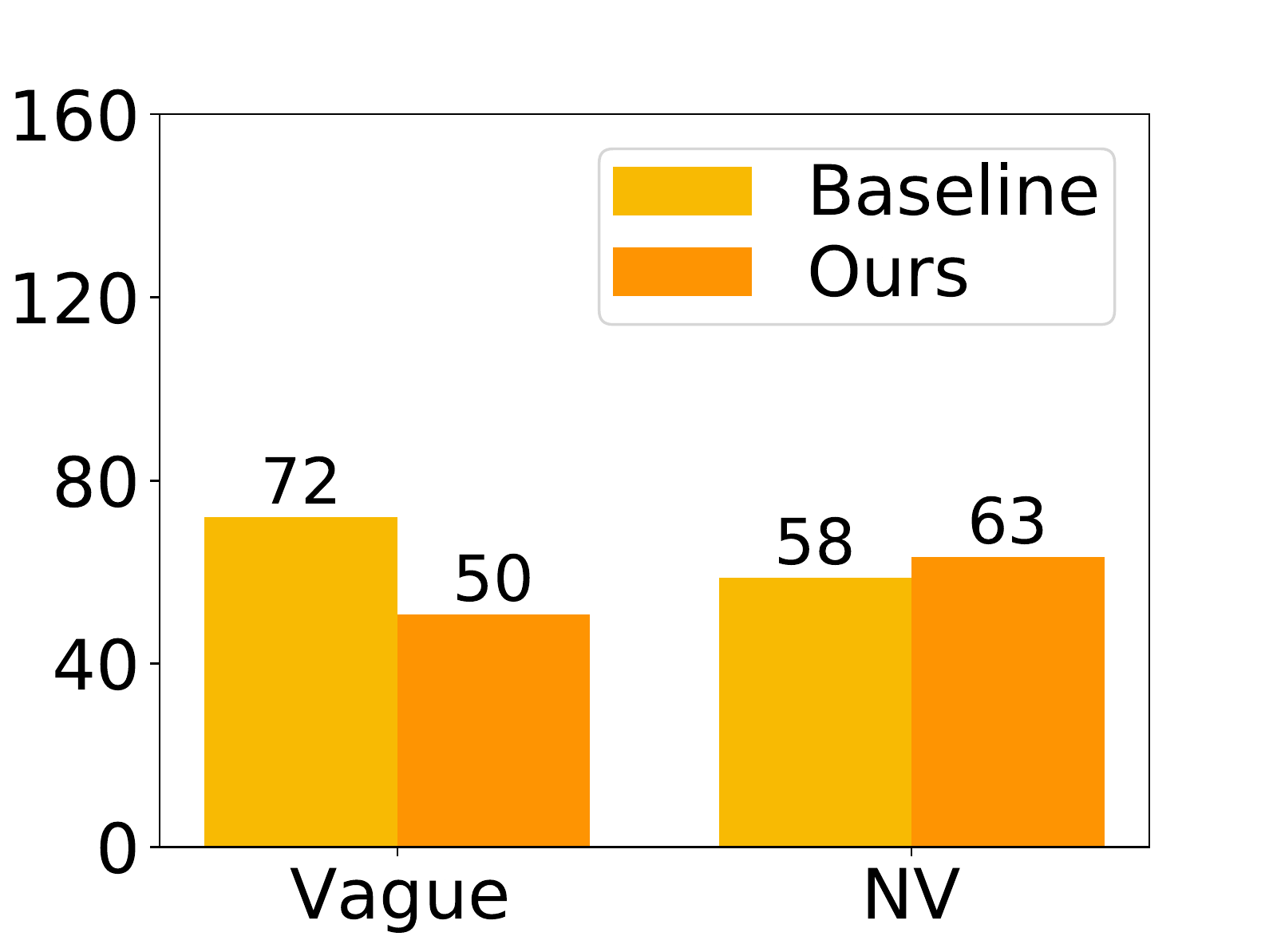}
       }

    \caption{Incorrect cases number of baseline (Enhanced-Baseline) and our model on both TB-Dense and MATRES.}
    \label{fig:vague_analyse}
\end{figure}

For classification-based methods, every label is treated as a one-hot vector, and models have to guess these vectors' meanings through training data. In contrast, we interpret every temporal relation into a logical expression of start and end time points, which clearly states the meaning of the relation. Among all the temporal relations between two events, \textit{Vague} is the most ambiguous one, because it does not describe a specific situation and it could correspond to various possibilities. Therefore, it is very hard for a model to summarize this relation's meaning from training data. We will show using logical expressions to make clear definitions, could benefit ambiguous relations, like \textit{Vague}.

We focus on the positive instances whose gold label is not \textit{Vague}, and Figure~\ref{fig:vague_analyse} shows the number of instances misclassified as relation \textit{Vague}, and the number of instances misclassified as others, which is denoted as NV. We can see that  \textit{Vague}-related errors make up the majority, which reflects the challenge posed by the ambiguity of \textit{Vague}. Comparing the performance of the baseline and ours, we see that the number of errors associated with \textit{Vague} decreases by 95 and 22 in TB-Dense and MATRES, respectively. This significant decrease indicates that by explicitly interpreting the meaning of \textit{Vague} using a logical expression, our approach can help the model better understand this relation and alleviate the confusion between this relation and others. There is a slight increase of errors not related to \textit{Vague} on both datasets. These  errors are mainly related to \textit{Before} and \textit{After}, whose meaning is not so ambiguous and thus may not benefit from our approach.

\subsection{Capability of Capturing Dependency}

\begin{table}[t]
\center
\setlength\tabcolsep{4pt}
\small
\begin{tabular}{@{}llllllll@{}}
\toprule
                                              &          & 1\% & 5\% & 10\%  & 20\%  & 30\%  & Avg \\ \midrule
\multirow{3}{*}{Mi-F1}                     & Base & 28.8 & 47.1 & 51.4 & 57.9 & 60.3 &     \\
                                              & Ours     & 29.2 & 50.8 & 56.5 & 60.6 & 62.8 &     \\
                                              & $\Delta$ & +0.4  & +3.7  & +5.1  & +2.7  & +2.5  & +2.9 \\ \midrule
\multirow{3}{*}{Ma-F1} & Base & 13.8 & 25.7 & 32.2 & 37.0 & 39.5 &     \\
\multicolumn{1}{c}{}                          & Ours     & 16.6 & 27.7 & 33.0 & 38.6 & 41.1 &     \\
\multicolumn{1}{c}{}                          & $\Delta$ & +2.8  & +2.0  & +0.8  & +1.6  & +1.6  & +1.8 \\ \bottomrule
\end{tabular}

\caption{Performance of baseline (Enhanced-Baseline) and our model in low-data scenarios. We use 1\%, 5\%, 10\%, 20\% and 30\% data in TB-Dense to train a BERT-Base based model. Mi-F1 and Ma-F1 represent Micro-F1 and Macro-F1, respectively.}
\label{table:fewshot}
\end{table}

Classification-based methods treat relations as independent labels and thus the instance of one relation could not help the model to understand other relations. Different from such methods, we represent all temporal relations in a unified framework and different relations could be connected via specific time point pairs. For example, \textit{Before} and \textit{Includes} in TB-Dense share similar relations between the start points of two events, which is $t^1_s \leq t^2_s$. Thanks to such connections, when a model meets an instance whose relation is \textit{Before}, the model could also learn something about \textit{Includes}. This enables the model to leverage relations with sufficient training data to aid in the understanding of relations whose training data is limited.

We show our method could improve the efficiency of data utilization by analyzing the performance in low-data scenarios.  Due to the unbalanced label distribution, relations like \textit{Includes} have very few training samples in low-data scenarios and thus it would be hard to learn by itself. We randomly sample 1\%, 5\%, 10\%, 20\%, and 30\% cases from TB-Dense, and Table~\ref{table:fewshot} shows the performance of the baseline and our method.

Overall, our method achieves a stable improvement compared to the baseline in all settings. On average, our method outperforms the baseline by 2.9\% and 1.8\% for micro-F1 and macro-F1, respectively. This shows that our method is capable of using data more effectively. 
As shown in Table~\ref{table:tre_compare},  our method improves 1.9\% micro-F1 compared to the baseline when trained with the whole TB-Dense, which is lower than the average improvement under low resources, indicating that our method has more potential in low resource scenarios. 
We note that in the scenario with the smallest amount of training data, i.e., setting 1\%, the difference of micro-F1 between ours and the baseline is relatively small. This is because, in this scenario, there are few instances corresponding to the relations \textit{Includes}, \textit{Is\_Included} and \textit{Equal}, and the baseline model directly degenerates into a 3-way classifier, only predicting \textit{Before}, \textit{After} and \textit{Vague}. As \textit{Before} and \textit{After} also account for most of the test sets, the baseline achieves a good micro-F1. Our  method, on the other hand, is capable of learning relations like \textit{Includes}, which has limited training samples, through relations with sufficient data, like \textit{Before}. The good comprehension of relations with limited data is demonstrated by the significant improvement on macro-F1, where our method outperforms the baseline by 2.8\%.

\subsection{Adaptation to Different Definitions}

\begin{table}[t]
    \centering
    \small
    \resizebox{0.48\textwidth}{!}{
    \begin{tabular}{c|c|c}
        \toprule
        Model& Normal & Transfer \\
        \midrule
        Baseline(Mapping1) & 81.9 $\pm$ 0.35 & 63.1 $\pm$ 1.1 \\
        Baseline(Mapping2) & 81.9 $\pm$ 0.35 & 64.3 $\pm$ 0.7 \\
        Ours & 82.6 $\pm$ 1.05& 70.4 $\pm$ 0.8\\
        \bottomrule
    \end{tabular}
    }
    
    \caption{Results of the transfer learning experiment. Normal indicates models are trained on MATRES and tested on MATRES. Transfer indicates models are trained on TB-Dense and tested on MATRES.}
    \label{table:transfer_learning}
\end{table}

One advantage of modeling the relation between time points instead of directly predicting the relation between events is that our method can be adapted to different task definitions. The relations in different task definitions, though, have different meanings, all of them could be interpreted using the relation between time points. For example, TB-Dense and MATRES have different relation definitions, but we could learn how to determine the relation between $t^1_s$ and $t^2_s$ from TB-Dense and then use this kind of time point relation to directly infer the temporal relations in MATRES. In other words, we only need to modify the logic expressions that map  the time point relations to the event relations, when the task definitions are changed, and do not have to train a new model from scratch.

The situation is quite different for methods that directly predict the relationships between events. This is because there is not a strict one-to-one mapping between different task definitions. One typical example is the relation \textit{Vague} in TB-Dense. It might indicate the relation between the start time of the two events is uncertain or the two events start in a determined order but it is hard to determine which event ends first. In this case,  the \textit{Vague} in TB-Dense may correspond to all four relations in MATRES.  Another example is that \textit{Includes} in TB-Dense indicates that the start time of event 1 is no later than event 2, which could be either the \textit{Before}, \textit{Equal}, or \textit{Vague} relation in MATRES.

We evaluate models' ability to adapt to different task definitions by training on TB-Dense and testing on MATRES. For our approach, the time point sorter of $t^1_s$ and $t^2_s$ trained on TB-Dense can be directly used to infer the relations in MATRES. And for the baseline model, it is necessary to map the relations in TB-Dense to the relations in MATRES. Firstly, \textit{Before}, \textit{After}, \textit{Simultaneous} and \textit{Vague} in TB-Dense are mapped to \textit{Before}, \textit{After}, \textit{Equal} and \textit{Vague} in MATRES, respectively. Then for the remaining two relations, \textit{Includes} and \textit{Is\_Included in}, we apply two different mappings, one is to map them both to \textit{Vague} in MATRES because we could not determine the specific start time relation, which is denoted as Mapping1. The other is to map \textit{Includes} to \textit{Before} and \textit{Is\_Included} to \textit{After}, considering the probability of two events starting simultaneously is small, and we denote this as Mapping2.

Table~\ref{table:transfer_learning} shows the average micro-F1 score and standard deviation of two models using RoBERTa-Large. We can see that, our model outperforms the baseline by 0.7\% F1 when trained and tested both on MATRES. As the training set changed from TB-Dense to MATRES, there is a significant increase in the gap between our model and baseline with both two mapping methods. We outperform Mapping1 by 7.3\% and outperform Mapping2 by 6.1\%, which shows our advantage in transfer learning. By representing all relations in a unified framework, our model bridges the gap between MATRES and TB-Dense, which demonstrates the strong generalization capability of our method.

\section{Event RE in LLMs}
\label{sec:event_re_in_llms}
\begin{figure}[h]
\center
  \includegraphics[width=0.45\textwidth]{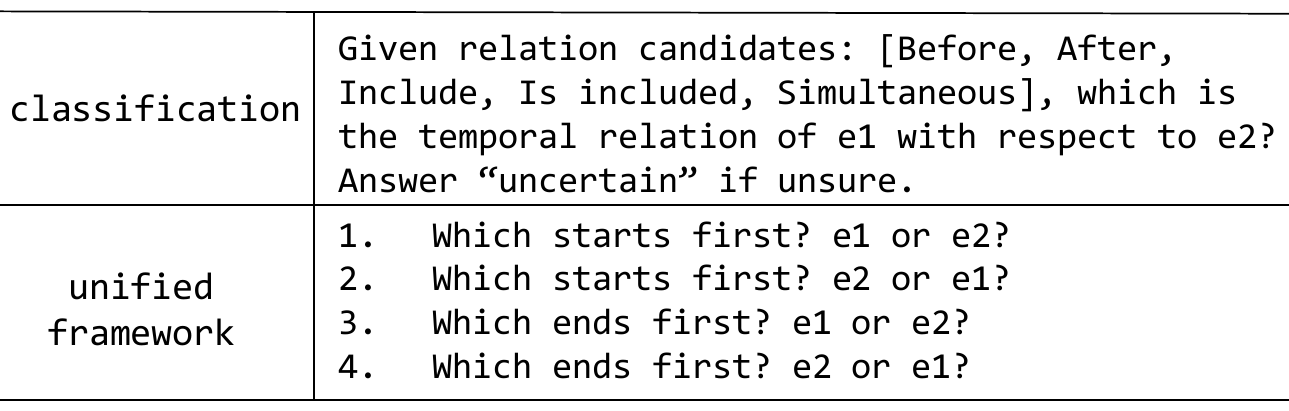}
    \caption{An illustration of the two kinds of prompts we used. The complete prompts can be found in Appendix~\ref{sec:app_experiment_details}. The unified framework asks the model to answer the four questions and deduce the final temporal relation based on the answers.}
    \label{fig:prompt_example}
\end{figure}

Large Language Models (LLMs), such as ChatGPT\footnote{https://chat.openai.com/}, have shown impressive performance in various tasks. In this section, we investigate the performance of LLMs in event temporal relation extraction and assess the value of our proposed unified framework in the era of LLMs.

We conduct experiments using gpt-3.5-turbo-0301\footnote{https://platform.openai.com/docs/models/gpt-3-5}, and Figure~\ref{fig:prompt_example} shows the prompts we used. The classification-based prompt lists all the candidate relations and requires the model to select one from them. If the list of relation candidates changes, the prompt should be modified accordingly and the old results will be useless.
Unlike the classification-based method, our unified framework is irrelevant to the candidate list. We ask the model to answer four questions and deduce the temporal relation based on the answers, just following what we do on Bert-based models. We explore variations of prompts and the detailed experimental settings can be found in Appendix~\ref{sec:app_experiment_details}. Table~\ref{tab:gpt-results} shows the results on TBDense, and below are our main findings:

\textbf{The order of candidates matters in the classification-based prompt.} The distribution of temporal relations is highly imbalanced in existing datasets, and we find that putting the majority relation, which is \textit{Before}, at the beginning of the candidate list will significantly affect the performance. For the vanilla classification-based prompt, we randomly sample an order of candidates for different cases. In contrast, \textit{+Before First Order} and \textit{+Before Last Order} use a fixed order, which put \textit{Before} at the beginning or end of the candidate list, respectively.\footnote{Please refer to Figure~\ref{fig:prompt_order} for details.} As shown in Tabel~\ref{tab:gpt-results}, compared with the other two orders, putting \textit{Before} at the beginning causes at least 2.7\% decline in F1. Further analysis shows that in this scenario, the model is more likely to predict \textit{Before}, making up to 55\% of all the predictions.

\textbf{Chain of thought (CoT) can improve accuracy.} Instead of directly generating the temporal relation between two events,  generate the reasoning procedure first and then deduce the answer could provide 0.5\% improvement in F1.

\textbf{Single word answer might not determine the direction of relations.} When the model is expected to return a single word, like \textit{Before} to represent the temporal relation between two events, it might mean \textit{$e_1$ before $e_2$}, but it could also mean \textit{$e_2$ before $e_1$}. This is a common phenomenon when the prompt does not explicitly mention the direction of relation, e.g., ``What is the temporal relation between $e_1$ and $e_2$``. This would lead to inferior performance, that the vanilla classification-based prompt outperforms the \textit{-Relation Direction Prompt} by 3.3\% in F1.

\textbf{A unified framework may help Large Language Models (LLMs).} As shown in Table~\ref{tab:gpt-results}, using our unified framework prompt could achieve 41.0\% F1, surpassing all classification-based variants, including the variant that incorporates self-consistency trick \cite{wang2022self}. 

\begin{table}[]
\small
\center
\begin{tabular}{llll}
\hline
                         & P     & R     & F1    \\ \hline
Classification-Based     & 28.7 & 48.9 & 36.1 \\
+ Before Fist Order      & 26.7 & 44.7 & 33.4 \\
+ Before Last Order      & 29.9 & 51.7 & 37.9 \\
- Relation Direction     & 29.8 & 36.4 & 32.8 \\
+ CoT                    & 31.5 & 43.9 & 36.6 \\
+ CoT + Self-Consistency & 33.8 & 45.5 & 38.7 \\ \hline
Unified Framework(Ours)  & 42.1 & 39.9 & 41.0 \\ \hline
\end{tabular}
\caption{Performance of ChatGPT on TBDense}
\label{tab:gpt-results}
\end{table}

\section{Related Work}

Earlier studies have proposed various definitions of the relations between two events, and all of them adopt Allen's interval representation. The 13 interval relations together with 1 \textit{Vague} relation form the basis elements for other relation definitions. TimeBank \cite{pustejovsky2003timebank} and TempEval-3 \cite{uzzaman2013semeval} directly use all of the 14 relations and then the researchers find that some relations are too fine-grained for both humans and models. Thus they simplify the ETRE task by aggregating some relations into a coarse one, e.g., \citet{verhagen2007semeval} merged all the overlap relations into a single relation, overlap. ISO-TimeML \cite{pustejovsky2010iso} pays attention to a special relation, \textit{contain}, which is composed of three interval relations where one interval is within the other. The focus on relation \textit{contain} influences many followers, like THYME \cite{styler-iv-etal-2014-temporal}, Richer \cite{o2016richer}, TB-Dense \cite{cassidy2014annotation} and MAVEN \cite{wang2022maven}. All these definitions, though differ in various aspects, can convert to intervals relations and thus could be interpreted using the endpoints of intervals. In other words, the different relations under these definitions could be represented in our unified framework. We just use two most widely used definitions, TB-Dense and MATRES, to evaluate our framework, and our framework can be applied to other definitions.

To solve the ETRE, previous efforts often regarded this task as a classification problem, and focus on learning better representations of event pairs, e.g, incorporating syntactic information \cite{meng-etal-2017-temporal,choubey-huang-2017-sequential,DBLP:journals/corr/abs-2104-09570} or discourse information \cite{mathur2021timers} into the encoder. Some studies also try to design auxiliary tasks, like event extraction \cite{DBLP:journals/corr/abs-2104-09570} or relative event time prediction \cite{wen-ji-2021-utilizing} to further enhance the encoder. Different from their work, we focus on helping the model understand temporal relations better after obtaining the event pair representation from the encoder, and thus our work is orthogonal to them. Recently, \citet{hwang-etal-2022-event} uses a box embedding to handle the asymmetric relationship between event pairs. However, the box embedding can only handle four types of relations, i.e., \textit{Before}, \textit{After}, \textit{Equal} and \textit{Vague}, and it cannot generalize to more complex relations, like \textit{Includes} in TB-Dense. To solve another task, \citet{cheng2020predicting} also consider start and end times separately. However, \citet{cheng2020predicting} directly uses a classifier to determine the relation between time points and cannot understand the relation \textit{Vague} well.

\section{Conclusion}
In this paper, we interpret temporal relations as a combination of  start and end time points of two events. Using this interpretation, we could not only explicitly convey temporal information to the model, but also represent relations of different task definitions in a unified framework. Our experimental results in TB-Dense and MATRES demonstrate the effectiveness of our proposed method, significantly outperforming previous state-of-the-art models in full data setting and providing large improvements on both few-shot and transfer-learning settings. In the future, we will investigate the potential of our approach in cross-document scenarios.

\section*{Acknowledgements}
This work is supported in part by National Key R\&D Program of China (No. 2020AAA0106600) and NSFC (62161160339). We would like to thank the anonymous reviewers for their helpful comments and suggestions; thank Weiye Chen for providing valuable comments. For any correspondence, please contact Yansong Feng.

\section*{Limitations}
Due to the limitation of dataset resources, we evaluate our unified model only with TB-Dense and MATRES. Although the experiment results show that our approach can significantly outperform state-of-the-art methods, we still need to experiment on more datasets with various kinds of temporal relations to further prove the generalization capability and robustness of our framework. 

\bibliography{custom}
\bibliographystyle{acl_natbib}
\newpage
\appendix

\section{Allen's Interval Relations}
\label{App:allen_relations}
\begin{figure}[h]
\center
  \includegraphics[width=0.45\textwidth]{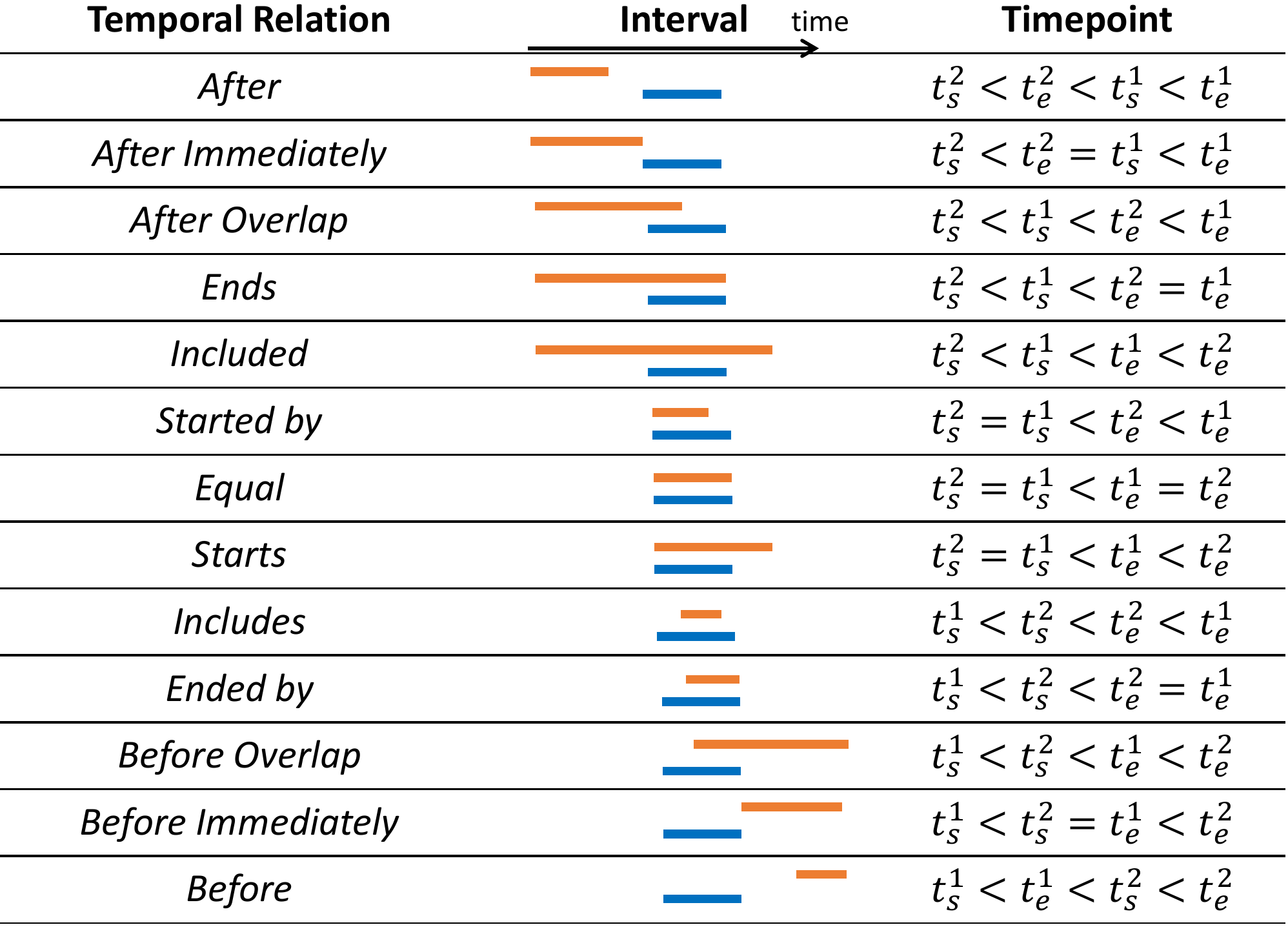}
    \caption{All 13 interval relations defined in \citet{allen1981interval}. \textcolor{myblue}{\rule[1.5pt]{0.5cm}{0.15em}} and \textcolor{myorange}{\rule[1.5pt]{0.5cm}{0.15em}} represent the intervals of event 1 and event 2 in the timeline. $t^*_s$ and $t^*_e$ represent the start and end time points of an event.}
    \label{fig:full_point2interval}
\end{figure}

\section{Rules of Symmetry}
\label{App:sym_rule}
\begin{table}[h]
    \centering
    \small
    \resizebox{0.48\textwidth}{!}{
    \begin{tabular}{c|c}
        \toprule
        \bf{Original Relation} & \bf{Symmetry Relation}\\
        \midrule
        A \textit{Before} B & B \textit{After} A \\
        A \textit{After} B & B \textit{Before} A \\
        A \textit{Include} B & B \textit{Is\_included} A \\
        A \textit{Is\_included} B & B\textit{ Include} A \\
        A \textit{Equal(Simultaneous)} B & B \textit{Equal(Simultaneous)} A \\
        A \textit{Vague} B & B \textit{Vague} A \\
        \bottomrule
    \end{tabular}
    }
    
    \caption{Symmetry rules between temporal relations}
    \label{table:rules_of_symmetry}
\end{table}

\section{MATRES Relations}
\label{App:matres-relations}
\begin{figure}[h]
\center
  \includegraphics[width=0.48\textwidth]{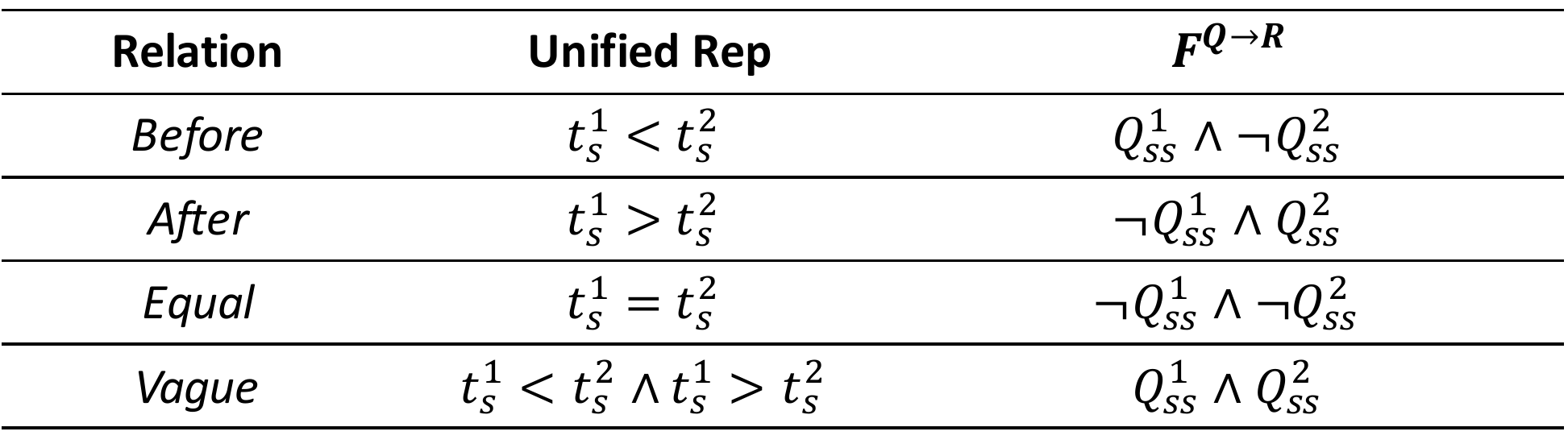}
    \caption{Relations in MATRES, and their unified representations and the logical expressions from Q to R($F^{Q\rightarrow}$)}
    \label{fig:matres-relations}
\end{figure}

\section{Implementation Details}
For fair comparisons with previous baseline methods, we use the pre-trained BERT-Base and RoBERTa-Large models for fine-tuning and optimize our model with AdamW. We optimize the parameters with grid search: training epoch $\in$ {1, 3, 5, 10}, learning rate $\in$ {2e-5, 1e-5}, training batch size 16, temperature in time point sorter $\in$ {1, 10}. The best hyperparameters for BERT-Base are (1, 2e-5, 10) and the best hyperparameters for RoBERTa-Large are (3, 1e-5, 10). We using one A40 GPU for training. 

\section{Experiment Details for LLMs}
\label{sec:app_experiment_details}

\label{App:prompt}

\begin{figure*}[h]
\center
  \includegraphics[width=1.0\textwidth]{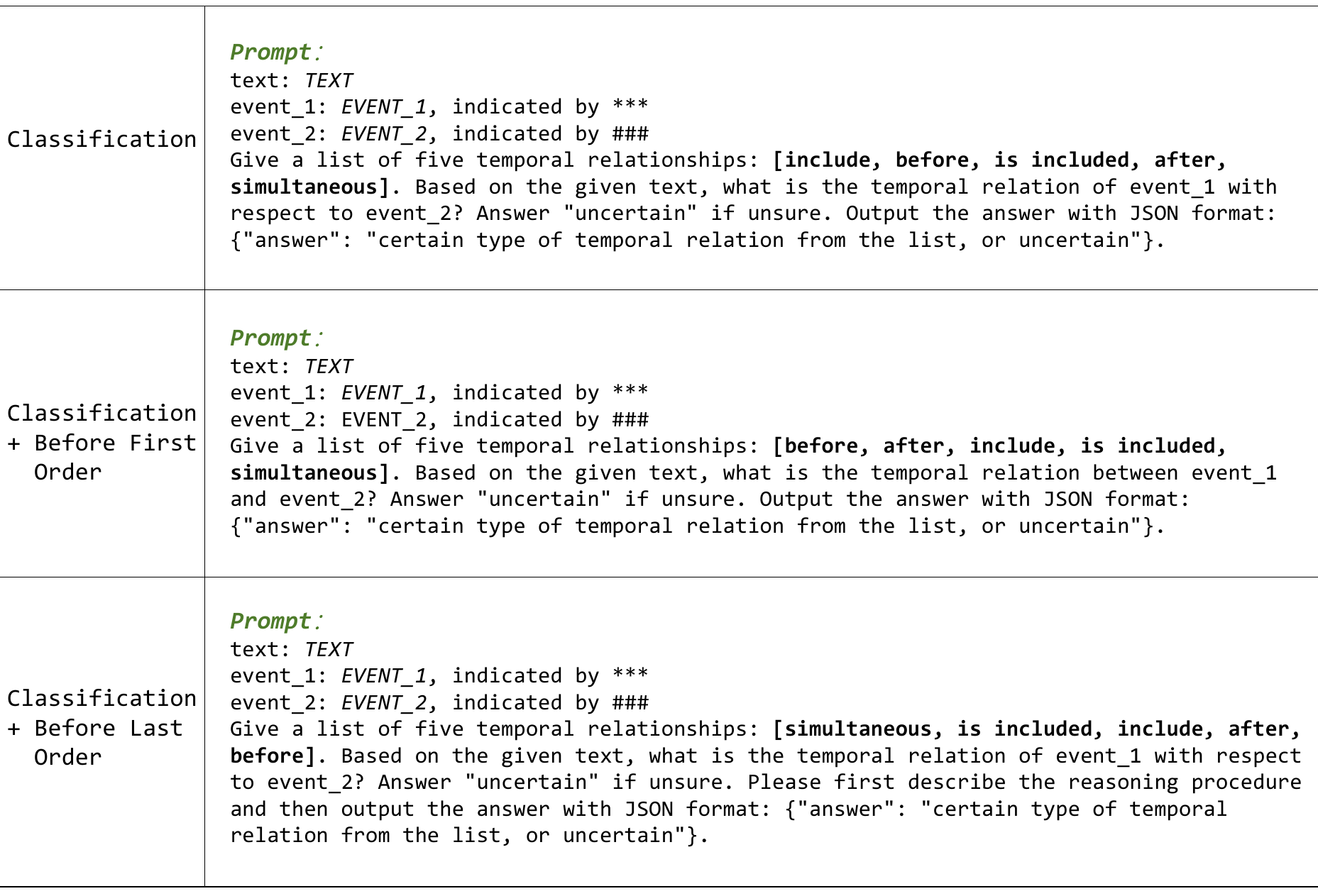}
    \caption{The details of the designed prompt. In Classification, the order of the temporal relationships is generated randomly. In Classification + Before First Order and Classification + Before Last Order, the order of the temporal relationships is fixed, based on the frequency of relationships in the test dataset. Classification + Before First Order put the most common relation \textit{Before} at the beginning of the list, while Classification + Before Last Order put \textit{Before} in the end.}
    \label{fig:prompt_order}
\end{figure*}
\begin{figure*}[h]
\center
  \includegraphics[width=1.0\textwidth]{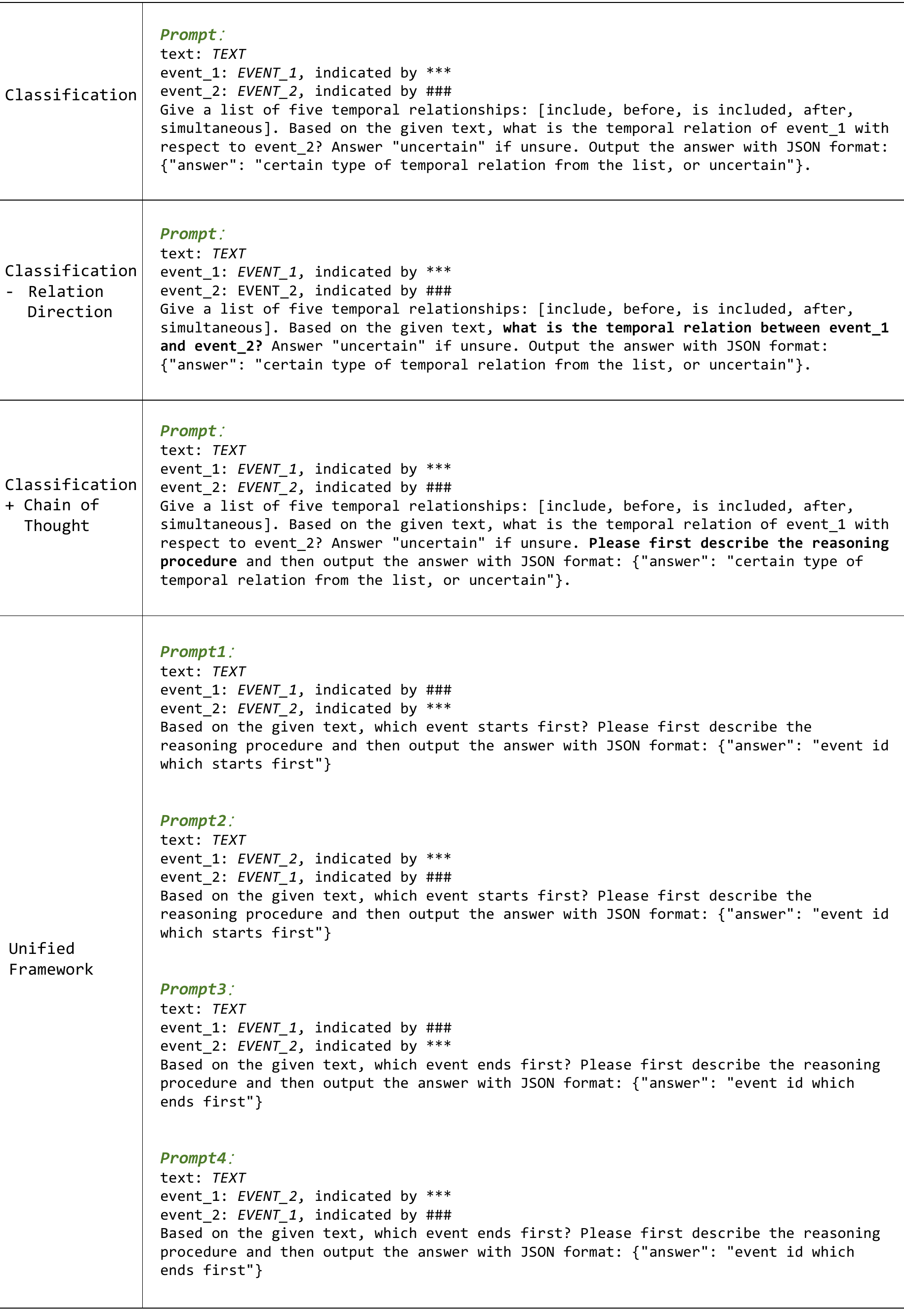}
    \caption{The details of the designed prompt. \textit{TEXT} represents the context containing event\_1 trigger \textit{EVENT\_1} and event\_2 trigger \textit{EVENT\_2}. The location of \textit{EVENT\_1} and \textit{EVENT\_2} in the \textit{TEXT} are emphasized by adding makers \#\#\# and *** in front of them respectively.}
    \label{fig:prompt}
\end{figure*}

\begin{table}[]
\center
\begin{tabular}{@{}p{1cm}p{1cm}c@{}}
\toprule
Pi         & Pj        & Time Point Relation\\ \midrule
$e_1$     & $e_2$    & before                 \\ \midrule
$e_1$     & $\phi$         & before                 \\ \midrule
$\phi$          & $e_2$    & before                 \\ \midrule
$e_2$     & $e_1$    & after                  \\ \midrule
$\phi$          & $e_1$    & after                  \\ \midrule
$e_2$     & $\phi$         & after                  \\ \midrule
\multicolumn{2}{l}{otherwise} &vague                  \\ \bottomrule
\end{tabular}
\caption{Mapping of the answers of prompts to the relation of time points. When Pi and Pj refer to Prompt1 and Prompt2 in Figure~\ref{fig:prompt}, we can deduce the relation of the start time point. And when Pi and Pj refer to Prompt3 and Prompt4 in Figure~\ref{fig:prompt}, we can deduce the relation of the end time point. $e_1$ and $e_2$ indicate the possible answer of LLMs: event\_1 and event\_2. $\phi$ means the output of LLMs is not in the label set \{event\_1, event\_2\}}
\label{tab:gpt_timepoint_mapping}

\end{table}

\begin{table}[]
\begin{tabular}{@{}ccc@{}}
\toprule
\multicolumn{1}{l}{Start Time} & \multicolumn{1}{l}{End Time} & Temporal Relation \\ \midrule
before                         & before                       & Before            \\
after                          & after                        & After             \\
before                         & after                        & Includes          \\
after                          & before                       & Included In       \\
\multicolumn{2}{c}{otherwise}                                 & Vague             \\ \bottomrule
\end{tabular}
\caption{Mapping from the relation of the start time point and the end time point to the final temporal relation between two events for ChatGPT.}
\label{tab:gpt_temporal_re_mapping}

\end{table}

Figure~\ref{fig:prompt} and \ref{fig:prompt_order} shows all the prompts we used in Section~\ref{sec:event_re_in_llms}. For the variants of classification-based prompts, we ask the model to directly output the temporal relation between two events. In Unified Framework, we design four prompts to first determine the relationship between start and end time points and then deduce the final temporal relation. 
Specifically, we ask LLMs which event starts first with \textit{Prompt1} and \textit{Prompt2} in Figure~\ref{fig:prompt}. If the results of the two prompts keep consistent, which means the answers are (event\_1, event\_2) or (event\_2, event\_1) for the two prompts respectively, we can determine the temporal relation of the start points between the two events. Otherwise, the temporal relation of the start points is set to \textit{Vague}. Sometimes, LLMs may generate answers not in the label set \{event\_1, event\_2\}. If both the answers for \textit{Prompt1} and \textit{Prompt2} are not in the label set, we regard the relation as \textit{Vague}. If only one answer for the two prompts is not in the label set, we determine the relation of start time points solely based on the other. We use the same rules to obtain the relation of the end time points based on the answers of \textit{Prompt3} and \textit{Prompt4}. Table~\ref{tab:gpt_timepoint_mapping} shows the mapping of the answers of prompts to the relation of start and end time points, and Table~\ref{tab:gpt_temporal_re_mapping} shows how to get the final temporal relation between the two events.

\end{document}